%% file: manuscript-amspreprint.tex
\pdfoutput=1

\IfFileExists{./prepreamble-amspreprint.sty}{\RequirePackage[packages,theorems,changes]{./prepreamble-amspreprint}}{}

\makeatletter
\IfFileExists{./scoop-latex/scoop-packages.sty}{\providecommand*{\input@path}{}\edef\input@path{{./scoop-latex/}\input@path}}{}
\makeatother
\PassOptionsToPackage{style=authoryear-comp,backend=biber}{biblatex}
\PassOptionsToPackage{commentmarkup = footnote}{changes}    
\PassOptionsToPackage{frozencache}{minted}    

\documentclass[english]{amsart}

\RequirePackage{biblatex}
\RequirePackage{ltxcmds}
\IfFileExists{./preamble-amspreprint.sty}{\RequirePackage[packages,theorems,changes]{./preamble-amspreprint}}{}

\usepackage{scoop-local}

\IfFileExists{./postpreamble-amspreprint.sty}{\RequirePackage[packages,theorems,changes]{./postpreamble-amspreprint}}{}

\makeatletter
\@ifpackageloaded{changes}{
\definechangesauthor[name = {Viktor Martinek}, color = {red!80!black}]{VM}
\definechangesauthor[name = {Ophelia Frotscher}, color = {blue!80!black}]{OF}
\definechangesauthor[name = {Markus Richter}, color = {orange!80!black}]{MR}
\definechangesauthor[name = {Roland Herzog}, color = {green!70!black}]{RH}
}{}
\makeatother        

\makeatletter
\@ifpackageloaded{hyperref}{%
	\hypersetup{
		pdftitle = {Introducing Thermodynamics-informed Symbolic Regression},
		pdfauthor = {Viktor Martinek, Ophelia Frotscher, Markus Richter, Roland Herzog},
		pdfkeywords = {symbolic regression, thermodynamics, equations of state},
		pdfcreator = {Created using the Scoop Template Engine version 1.2.0.}
	}
}{
	\pdfinfo{
		/Title (Introducing Thermodynamics-informed Symbolic Regression)
		/Author (Viktor Martinek, Ophelia Frotscher, Markus Richter, Roland Herzog)
		/Subject ()
		/Keywords (symbolic regression, thermodynamics, equations of state)
		/Creator (Created using the Scoop Template Engine version 1.2.0.)
	}
}
\makeatother

\title[TiSR: Thermodynamic Equations of State Development]{Introducing Thermodynamics-informed Symbolic Regression \\ A Tool for Thermodynamic Equations of State Development}

\author{Viktor Martinek}
\address[V. Martinek]{Interdisciplinary Center for Scientific Computing, Heidelberg University, 69120 Heidelberg, Germany}
\email{viktor.martinek@iwr.uni-heidelberg.de}

\author{Ophelia Frotscher}
\address[O. Frotscher]{Technische Universität Chemnitz, Faculty of Mechanical Engineering, Applied Thermodynamics, 09107 Chemnitz, Germany}
\email{ophelia.frotscher@mb.tu-chemnitz.de}

\author{Markus Richter}
\address[M. Richter]{Technische Universität Chemnitz, Faculty of Mechanical Engineering, Applied Thermodynamics, 09107 Chemnitz, Germany}
\email{m.richter@mb.tu-chemnitz.de}

\author{Roland Herzog}
\address[R. Herzog]{Interdisciplinary Center for Scientific Computing, Heidelberg University, 69120 Heidelberg, Germany}
\email{roland.herzog@iwr.uni-heidelberg.de}

\thanks{This work was funded by the Deutsche Forschungsgemeinschaft (DFG, German Research Foundation) -- HE~6077/14-1 and RI~2482/10-1 -- within the Priority Programme \enquote{SPP 2331: Machine Learning in Chemical Engineering}. }

\date{\today}

\dedicatory{}

\begin{document}

\begin{abstract}
\input{abstract.tex}
\end{abstract}


\makeatletter
\ltx@ifpackageloaded{hyperref}{%
\subjclass[2010]{}
}{%
\subjclass[2010]{}
}
\makeatother

\maketitle

\input{content.tex}

\appendix

\printbibliography

\end{document}

%% file: abstract.tex
Thermodynamic \ac{EOS} are essential for many industries as well as in academia.
Even leaving aside the expensive and extensive measurement campaigns required for the data acquisition, the development of \ac{EOS} is an intensely time-consuming process, which does often still heavily rely on expert knowledge and iterative fine-tuning.

To improve upon and accelerate the \ac{EOS} development process, we introduce \ac{TiSR}, a \ac{SR} tool aimed at thermodynamic \ac{EOS} modeling.
\ac{TiSR} is already a capable \ac{SR} tool, which was used in the research of \cite{FrotscherMartinekFingerhutYangVrabecHerzogRichter:2023:1}.
It aims to combine an \ac{SR} base with the extensions required to work with often strongly scattered experimental data, different residual pre- and post-processing options, and additional features required to consider thermodynamic \ac{EOS} development.

Although \ac{TiSR} is not ready for end users yet, this paper is intended to report on its current state, showcase the progress, and discuss (distant and not so distant) future directions.
\ac{TiSR} is available at \url{https://github.com/scoop-group/TiSR} and can be cited as \cite{Martinek:2023:1}.

%% file: content.tex
\section{Introduction}
\label{section:introduction}

Accurate knowledge of the thermodynamic properties of pure fluids and mixtures is crucial for many scientific and technical tasks.
In research, thermodynamic property data are required, among other things, for the investigation of physical relationships and for the corresponding development of models.
For industry, property dat are an essential basis for the design of processes and equipment, whereby complex energy and process engineering plants can be developed and optimized using process simulation.
However, the quality of such simulations strongly depends on the accuracy of the available thermodynamic property data.
For the calculation of such data, \ac{EOS} are used.

The development of \ac{EOS} is a sophisticated and time-consuming process.
To address both, the measurement and the modeling, we recently proposed a new combined measurement and modeling procedure, see \cite{FrotscherMartinekFingerhutYangVrabecHerzogRichter:2023:1}.
There it is shown that using \ac{OED}, \ac{SR}, and hybrid data acquisition can significantly reduce measurement and development expenses for \ac{EOS}.
For this purpose, \ac{TiSR} was utilized, the newly developed \ac{SR} tool aimed at thermodynamic \ac{EOS} modeling.
It is an integral part of our proposed combined measurement and modeling procedure, as well as a post-processing tool.

There are many forms of thermodynamic \ac{EOS} for different purposes and with varying accuracy.
However, only very little has been published on improving functional form of thermodynamic models, \eg, of multi-parameter fundamental \ac{EOS}, in the recent decades.
\ac{SR} is well poised to change this and discover new model forms, especially for much simpler than fundamental \ac{EOS}.

As mentioned above, even with the measurement expenses left aside, the modeling process of thermodynamic \ac{EOS} itself is a time-consuming process, that does still rely on \enquote{expert experience and intuition} and iterative fine-tuning.
To remedy this, we are working on, among other things, the combination of an \ac{SR} algorithm with the additional features required for often scattered experimental data, the specific extensions for the thermodynamic \ac{EOS} development as well as the formalization and enforcement of some of the \enquote{expert experience and intuition} and especially thermodynamic constraints.
\ac{TiSR} is work-in-progress and does not yet have all mentioned features, nor is it ready for end users.
However, we are working with further measurement and modeling experts to incorporate all of the above and more.
Nevertheless, \ac{TiSR} is available at \url{https://github.com/scoop-group/TiSR} and can be cited as \cite{Martinek:2023:1}.

\section{Design Philosophy}
\label{section:design_phil}

The current version of \ac{TiSR} is not aiming to be a state-of-the-art general purpose \ac{SR} tool.
Its symbolic regression basis is relatively simple but flexible.
Its goal is to introduce the physics- and thermodynamics-specific extensions to create a tool for the thermodynamic \ac{EOS} development.
At a later stage, these adaptions and extensions may be applied to other \ac{SR} algorithms.

The \ac{SR} library \symbolicregression by \cite{Cranmer:2020:1} was used as a reference and starting point for \ac{TiSR}.
However, numerous adaptions were made at many levels, sacrificing general performance to gain flexibility, extensibility, and simplicity.
As a rule of thumb and to give a general idea, when given the choice, the pragmatic approach of using half the code is used, even if it reduces performance by $\approx~20\%$.
For our work in progress tool \ac{TiSR}, this approach allows us to do fast prototyping and extend the core functionality by several features, some of them thermodynamics-specific.
Performance optimization, which is a necessity for genetic algorithms, will come at a later stage, when most or all of the features we, and others utilizing \ac{TiSR}, envision and require are implemented and tested.
Nevertheless, the current performance is sufficient for academic purposes.

\section{Overview and Features}
\label{section:features}

\ac{TiSR} is written in the programming language Julia developed by \cite{BezansonEdelmanKarpinskiShah:2017:1}.
It utilizes a genetic programming (see \cite{Koza:1994:1}) algorithm, \ie, a modified version of NSGA-II (see \cite{DebPratapAgarwalMeyarivan:2002:1}) with an island model population structure (see \cite{GorgesSchleuter:1991:1}).

\subsection{Overview}
\label{subsection:overview}

\ac{TiSR}'s main loop consists of expression mutation (\cref{subsection:mutation}), individual instantiation (described next), and selection (\cref{subsection:selection}).
Every iteration of this main loop constitutes a generation of the genetic algorithm.

Each individual contains one expression and a number of related attributes, which are determined in the \enquote{individual instantiation}, which is divided in the following steps:

\begin{enumeratearabic}
	\item
		unnecessary parameter removal
		\begin{itemize}
			\item
				for example: parameter + parameter $\rightarrow$ parameter, function(parameter) $\rightarrow$ parameter, \ldots
			\item
				values of parameters are not adjusted, as parameters are identified after this step
		\end{itemize}
	\item
		randomly trimming expressions that exceed the size limit
	\item
		reordering of operands of $+$ and $\cdot$
		\begin{itemize}
			\item
				for example: $v_1 + 1 \rightarrow 1 + v_1$, $3 \cdot \cos(v_2 + 1) \rightarrow 3 \cdot \cos(1 + v_2)$
			\item
				reorder according to following rules ($<$ means before):
				\\
				$\text{parameter} < \text{variable} < \text{unary operator} < \text{binary operator}$
		\end{itemize}
	\item
		grammar checking (described below)
	\item
		parameter identification (\cref{subsection:eval})
		\begin{enumerate}
			\item
				calculate residual-related measures (see \cref{table:individual_attributes})
			\item
				calculate constraint violations (coming soon)
		\end{enumerate}
	\item
		singularity prevention (coming soon)
	\item
		determination of attributes unrelated to the residual (see \cref{table:individual_attributes})
\end{enumeratearabic}

At the \enquote{grammar checking}, \enquote{parameter identification} and the \enquote{singularity prevention} steps, individuals may be deemed invalid, resulting in their termination and removal.
The use of grammar may increase the algorithm's performance by filtering out individuals before parameter identification.
Currently, two grammar options are available.
The user may prohibit certain operator compositions, \eg, $\cos(\cos(x))$ or $\exp(\log(x))$.
These are also enforced during the random creation of expressions.
The second currently implemented grammar option prohibits parameters in exponents, \ie, $(x + 1)^3$ would be allowed but $3^{(x + 1)}$ would not.
The latter grammar is not enforced at the \enquote{grammar checking} step above, but rather throughout \ac{TiSR} in the individual creation and mutations.
We plan to introduce more grammar options in the future.

The attributes of each individual, their brief description, and whether the attributes are related to the residual can be seen in \cref{table:individual_attributes}.
Currently, the ability to add custom attributes by the user is not implemented but will be added in the future.

\begin{table}[h]
	\centering
	\caption{Individual attribute names, a short description, and whether they are related to the residual.}
	\begin{tabular}{p{0.22\textwidth} | p{0.1\textwidth} | p{0.65\textwidth}}
		Measure                                  & residual-related   & Description \\ \hline
		\mintinline{julia}{ms_processed_e}       & yes                & measure, which is minimized in the parameter identification (discussed in more detail \cref{subsection:eval})
		\\
		\mintinline{julia}{mse}                  & yes                & mean squared error
		\\
		\mintinline{julia}{mae}                  & yes                & mean absolute error
		\\
		\mintinline{julia}{max_ae}               & yes                & maximum absolute error
		\\
		\mintinline{julia}{minus_r2}             & yes                & negative coefficient of determination $-r^2$
		\\
		\mintinline{julia}{mare}                 & yes                & mean relative error
		\\
		\mintinline{julia}{q75_are}              & yes                & $75\%$ percentile of the mean relative error
		\\
		\mintinline{julia}{max_are}              & yes                & maximum absolute relative error
		\\
		\mintinline{julia}{compl}                & no                 & complexity, \ie, number of variables, parameters, and operations
		\\
		\mintinline{julia}{recursive_compl}      & no                 & recursive complexity as proposed by \cite{KommendaBehamAffenzellerKronberger:2015:1} (with some minor extensions to, among other things, incorporate the power-operator)
		\\
		\mintinline{julia}{n_params}             & no                 & number of parameters
		\\
		\mintinline{julia}{age}                  & no                 & number of generations since creation of the expression
		\\
		\mintinline{julia}{valid}                & no                 & whether the individual is valid, as determined by \enquote{check grammar}, \enquote{identify parameters}, and the \enquote{prevent singularities}
	\end{tabular}
	\label{table:individual_attributes}
\end{table}

For parallelization, we currently employ multithreading on the \enquote{individual instantiation} step which includes parameter identification.
As the parameter identification is by far the most expensive step, parallelizing the complete generational loop offers only a small additional performance benefit.

\subsection{Population Structure}
\label{subsection:migration}

The island population model we use maintains several subpopulations which evolve separately.
The islands are arranged in a static ring topology (see \cite{LinPunchGoodman:1994:1}).
At a user defined generational interval, a random emigration island and a random direct neighbor, acting as immigration island, are chosen.
The migrating individual is chosen randomly from the emigration island's population, and it is copied (not moved) to the immigration island's population.

\subsection{Selection}
\label{subsection:selection}

The selection of individuals for the next generation is performed by non-dominated sort (Pareto optimal), or tournament selection (see \cite{Brindle:1980:1}), or both.
If both are used, the Pareto optimal selection is performed first, before tournament selection is performed on the remaining individuals.
The ratio to which the two selections are performed can be modified in the range $[0,1]$.

The selection objectives for both selection modes can be set individually.
Any of the attributes of the individuals, and any number of them, can be chosen (except for \mintinline{julia}{valid}) (see \cref{table:individual_attributes} for currently implemented attributes).

\subsection{Creating and Mutating Expressions}
\label{subsection:mutation}

In the first generation, instead of mutating expressions, new ones are at random.
The user may provide starting expressions, which allows to either incorporate domain knowledge or resume another run.
The currently implemented types of mutations are listed in \cref{table:mutations}.

\begin{table}[h]
	\centering
    \caption{Types of mutations currently implemented in \ac{TiSR}. The mutations with \enquote{deeper than $2$} are only chosen if the associated binary tree of the expression is deeper than~$2$.}
		\begin{tabular}{p{0.37\textwidth} | p{0.08\textwidth} | p{0.48\textwidth}}
			mutation                                       & deeper than $2$ & description
			\\ \hline
			\mintinline{julia}{insert_mutation!}           & no              & insert a random expression snippet at a random place
			\\
			\mintinline{julia}{point_mutation!}            & no              & choose a random element and exchange it with an equivalent one (variable $\rightarrow$ variable, binary operation $\rightarrow$ binary operation, \ldots)
			\\
			\mintinline{julia}{addterm_mutation!}          & no              & add a random term to the expression
			\\
			\mintinline{julia}{hoist_mutation!}            & yes             & choose random operator and remove it ($\cos(x) \rightarrow x$, $x + y \rightarrow x$, \ldots)
			\\
			\mintinline{julia}{innergrow_mutation!}        & yes             & select a random part of the expression and replace it with another part of the same expression (creating a new directed acyclic graph connection)
			\\
			\mintinline{julia}{subtree_mutation!}          & yes             & choose random operator and exchange it with a random expression snippet
			\\
			\mintinline{julia}{drastic_simplify!}          & yes             & remove all parameters in $+$, $-$, and $*$ operations which are smaller than set tolerance, and simplify accordingly ($x + 0.00001$ $\rightarrow$ $x$, $x + y \cdot 0.00001$ $\rightarrow$ $x$)
			\\
			\mintinline{julia}{simplify_w_symbolic_utils!} & yes             & simplify the expression using the \symbolicutils \cite{GowdaMaProtter:2023:1} package
			\\
			\mintinline{julia}{crossover_mutation!}        & yes             & randomly combine two individuals
		\end{tabular}
	\label{table:mutations}
\end{table}

In some cases, simplifications may lead to less desirable expressions in terms of the selection objectives.
Therefore, apart from the very basic simplifications to remove unnecessary parameters mentioned in \cref{subsection:overview}, we choose to perform the more complex simplification using the \symbolicutils package (see \cite{GowdaMaProtter:2023:1}) as mutations.

The \mintinline{julia}{drastic_simplify!} mutation removes parameters which are smaller than a set value in case they appear in an addition or subtraction.
In the context of multiplication, the complete term affected by the small parameter is removed.
This simplification helps guide the expression search, but it is especially powerful in combination with \lasso (least absolute shrinkage and selection operator) regression.
In \lasso regression, an $\ell_1$-norm of the parameter vector is added to the squared residual norm as a regularization term, which incentives potentially zero parameter values.
Currently, \ac{TiSR} does not support \lasso regularization (see \cite{Tibshirani:1996:1}) in the parameter estimation, but it will soon.

\subsection{Evaluation and Parameter Identification}
\label{subsection:eval}

For the evaluation of candidate expressions, the power, logarithm, and division operators are protected to allow their direct use, rather than necessitating the use of implicit domain restrictions, \eg, \mintinline{julia}{abs(x)^y}.
This protection is implemented by checking the operands and preventing the evaluation, if they are outside the valid domains for the respective operators.
These individuals are deemed invalid and removed from the population.
One noticeable benefit of this approach is that it makes simplifications of expressions in many cases easier.
For example, \mintinline{julia}{x * abs(x)^2} cannot be simplified without assuming $x > 0$ or $x < 0$.

The prediction of the expressions may be processed before the residual is calculated.
This can be used to search for parts of an expression, while presupposing the remainder.

Usually, \ac{SR} searches for an expression \mintinline{julia}{f(X)} which satisfies \mintinline{julia}{y = f(X)} for the given data \mintinline{julia}{X} and \mintinline{julia}{y}.
In this case, the residual vector is calculated using \mintinline{julia}{y - f(X)}.
However, in some cases, parts of the expression may be known.
If, for example, we search an expression \mintinline{julia}{f(X)} and presuppose that \mintinline{julia}{y = exp(f(X))} holds, we could rearrange the expression as \mintinline{julia}{log(y) = f(X)} and search for the \mintinline{julia}{f} part of the expression directly.
This however, does change the residual to \mintinline{julia}{log(y) - f(X)} and thus the minimization objective, which may lead to inferior results.
In other cases, the presupposed expression parts may not be possible to rearrange.
The expression \mintinline{julia}{y = f(X)^2 + exp(f(X))} cannot be solved for \mintinline{julia}{f(X)}.
In \ac{TiSR} it is possible to define \mintinline{julia}{pre_residual_processing!} function, which is applied to the output of the expression \mintinline{julia}{f(X)} before residual is calculated.
This allows to implement the transformations of the first or the second example above.
For the second example, for the evaluation of each expression \mintinline{julia}{f(X)} proposed by \ac{TiSR}, first, the output of the expression is calculated.
The \mintinline{julia}{pre_residual_processing!} function performs \mintinline{julia}{y_pred = f(X)^2 + exp(f(X))} to calculate the prediction of the proposed expression, and the residual is calculated by \mintinline{julia}{y - y_pred}.
All the residual-related measures listed in \cref{table:individual_attributes} are, of course, affected by this function.

For the parameter identification step, \ac{TiSR} employs a Levenberg-Marquardt algorithm (see \cite{Levenberg:1944:1, Marquardt:1963:1}).
We use a modified version of the implementation of \cite{WhiteMogensenJohnson:2022:1}.
The objective of the optimizer is to minimize the \mintinline{julia}{ms_processed_e} measure, which stands for mean squared processed error and has several user-exposed processing options.

By default, \mintinline{julia}{ms_processed_e} is equal to the \mintinline{julia}{mse}.
One of the customization options for the \mintinline{julia}{ms_processed_e} is to provide weights for the residuals to, \eg, minimize the relative rather than the absolute deviation, incorporate uncertainties, or both.
To process the residuals in other, possibly more complex ways, a custom function may be provided by the user.

To mitigate overfitting and improve generalization performance, early stopping (see \cite{MorganBourlard:1989:1}) is employed.
In early stopping, the parameter identification is conducted for a fraction of the data, while the residual norm is also calculated for the remainder.
One method to perform early stopping, as described in \cite{Prechelt:1998:1}, is to terminate the parameter estimation as soon as the residual norm for the remaining data increases monotonically for a number of iterations.
Apart from reducing overfitting, early stopping provides two added performance advantages.
First, the parameter identification, which is currently by far the most expensive part of the algorithm, may be stopped after fewer iterations for candidate expressions which do not appear to capture behavior underlying the data.
This decreases the performance penalty of allowing a larger maximum number of iterations during fitting while retaining its benefits.
Second, the Jacobian for the Levenberg-Marquardt algorithm is only calculated for a fraction of the data, which increases the performance for large data sets.

\section{Conclusion and Outlook}
\label{section:conclusion}

We introduced and provided a first overview of \ac{TiSR}.
Although it is work-in-progress, \ac{TiSR} is already a capable \ac{SR} tool, \eg, for \ac{EOS} modeling, with a genetic programming basis and similar capabilities as comparable tools.
Notable differences to other \ac{SR} algorithms in this combination are, among other things, the following:

\begin{itemize}
	\item 
		protected evaluation of power and similar functions, avoiding the need for constructions such as \mintinline{julia}{abs(x)^y}
		\begin{itemize}
			\item 
				improved simplification possibilities and more sensible expressions
		\end{itemize}
	\item 
		many residual pre- and post-processing options
		\begin{itemize}
			\item 
				weighting
			\item 
				custom post-processing
			\item 
				custom processing of the prediction before residual calculation to, \ie, search for sub-expressions
		\end{itemize}
	\item 
		removal of expression parts, which do not contribute much
		\begin{itemize}
			\item 
				targeted simplification
		\end{itemize}
	\item 
		early stopping
		\begin{itemize}
			\item 
				mitigate overfitting and improve generalization performance
			\item 
				improve performance by stopping earlier
		\end{itemize}
\end{itemize}

For the future, the most notable feature of \ac{TiSR} is its flexibility and extensibility, which allows us to implement major changes comparably easily and fast.

Many of the crucial features for \ac{EOS} development are planned to be implemented in the near future or are currently being implemented.
Our plans include:

\begin{itemize}
	\item 
		constrained fitting and constraint violation as additional selection objective
	\item 
		prevent singularities
	\item 
		support for factor variables as introduced by \cite{KommendaBehamAffenzellerKronberger:2015:1}, which allow the incorporation of nominal variables
	\item 
		support for the development of \ac{EOS} formulated in terms of the Helmholtz energy
	\item 
		\lasso regularization for better simplification
	\item 
		more grammar
	\item 
		experiment with other \ac{SR} algorithms at the basis
\end{itemize}